\documentclass[conference]{IEEEtran}
\IEEEoverridecommandlockouts
\usepackage[noadjust]{cite}
\usepackage{amsmath,amssymb,amsfonts}
\usepackage{algorithmic}
\usepackage{graphicx}
\usepackage{textcomp}
\usepackage{xcolor}
\usepackage{graphicx}
\usepackage{balance}
\usepackage[T1]{fontenc}
\usepackage{tikz}
\usepackage[bottom]{footmisc} 
\usepackage{upgreek}
\usepackage{bm}

\usepackage{float}
\usepackage{hyperref}
\def\BibTeX{{\rm B\kern-.05em{\sc i\kern-.025em b}\kern-.08em
    T\kern-.1667em\lower.7ex\hbox{E}\kern-.125emX}}
\begin{document}
\bstctlcite{IEEEexample:BSTcontrol}

\title{Single-Controller Teleoperation and Multimodal Dataset Generation for a Mobile Manipulator}

\author{Fardien Azizi$^{*1}$, Sitor Sla$^{*1}$, Christiaan Peppelman$^{*1}$, Dion van Santen$^{*1}$,\\ Andrei-Carlo Papuc$^{1}$, Khaled Mustafa$^{1}$, Diego Martinez-Baselga$^{1,2}$, Javier Alonso-Mora$^{1}$
\thanks{
 $^{*}$ Denotes equal contribution.}
\thanks{
 $^{1}$ The authors are with the Dept. of Cognitive Robotics, TU Delft, 2628 CD Delft, The Netherlands. (\texttt{\small \{f.azizi, s.j.n.sla, c.g.peppelman,\allowbreak d.e.l.vansanten\}@student.tudelft.nl, \{a.c.papuc, k.a.mustafa, d.martinezbaselga, j.alonsomora\}@tudelft.nl}).}
 \thanks{
 $^{2}$ The authors are with RoPeRt Group, DIIS-I3A, University of Zaragoza, 50018 Zaragoza, Spain. (\texttt{\small \{diegomartinez\}@unizar.es}).}%
\thanks{This work was supported by grants PID2022-139615OB-I00 and AIA2025-163563-C31 funded by MICIU/AEI/10.13039/501100011033 and ERDF/EU, from the Aragon Government/FSE-T45\_26R, and by the Dutch Research Council NWO-NWA, within the “Acting under uncertainty” (ACT) project (Grant No. NWA.1292.19.298), and by the Office of Naval Research Global (ONRG) under Grant N62909-25-12027 (Project SECURE). Views and opinions expressed are however those of the author(s) only and do not necessarily reflect those of the European Union or the European Research Council Executive Agency. Neither the European Union nor the granting authority can be held responsible for them.}%
}

\maketitle

\begin{abstract}
Robot learning systems require synchronized, high-quality demonstrations that combine visual, proprioceptive, and action data. However, collecting such datasets with mobile manipulators is often slowed down by fragmented teleoperation interfaces and manual post-processing workflows. This paper presents an integrated framework for teleoperated multimodal data collection on a Clearpath Dingo-O mobile base equipped with a Kinova Gen3 Lite manipulator and an Intel RealSense RGB-D camera. The framework combines single-controller operation of the base, arm, and recording process with an automated ROS-based pipeline for data capture, image-quality filtering, temporal synchronization, visualization, and export to machine-learning-ready Zarr datasets. We validate the system on a pick-and-place mobile-manipulation task in which the robot transports a bottle between two surfaces. The results show that the unified controller supports complete task execution and recording without external commands, while the data pipeline produces synchronized datasets for short demonstrations and identifies corrupted visual frames. The proposed framework provides a practical basis for producing structured demonstrations for robot-learning research. Code and videos of the experiments can be found at this \href{https://github.com/sjnsla/SCT-MMDC}{repo}.
\end{abstract}

\section{Introduction}

\begin{figure}[t]
    \centering
    \includegraphics[trim=0 0 0 600,clip,width=0.75\linewidth]{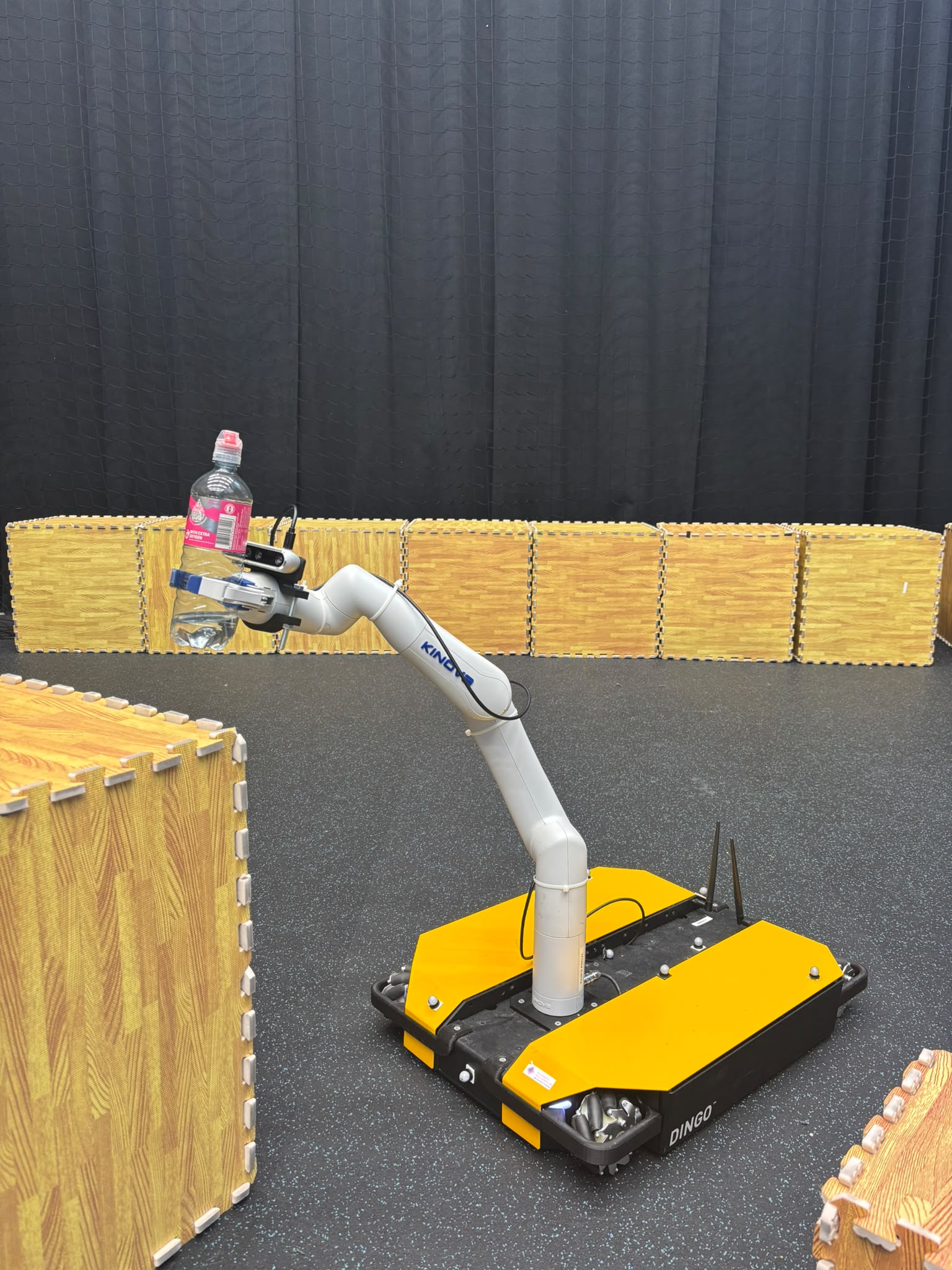}
    \caption{Mobile manipulator interacting with a bottle.}
    \label{fig:teaser}
\end{figure}

Mobile manipulators, such as the one in Fig.~\ref{fig:teaser}, combine a mobile base with a robotic arm to interact with objects across large and dynamic workspaces. Achieving reliable autonomy remains difficult because the robot must coordinate many degrees of freedom in partially known or unstructured environments. Deep learning offers a promising path toward learning such perception and control behaviors from data \cite{pierson2017deep}.

The importance of large and diverse robot datasets has become increasingly apparent as robot-learning models have grown in scale. RoboNet demonstrated the potential of combining interaction data from different robotic platforms \cite{dasari2020robonet}, while BridgeData showed how variation across tasks, objects, and environments can improve generalization \cite{walke2023bridgedata}. RT-1 similarly demonstrated that high-capacity policies can benefit from large collections of real-robot trajectories spanning many tasks \cite{brohan2023rt}. Open X-Embodiment extended this direction by combining data from multiple institutions and robot embodiments into standardized representations for cross-robot learning \cite{openx}. Together, these works indicate that the performance and generalization of robot-learning systems depend not only on the learning method, but also on the quantity, diversity, consistency, and quality of the available data.

Teleoperation is the primary mechanism for producing such data, since a human operator can generate task-relevant behavior while the robot records the corresponding observations and actions. However, turning teleoperation into a reliable data-acquisition pipeline raises four distinct challenges: demonstrations are costly and slow to collect, the intuitiveness of the control interface directly affects both collection speed and data quality, mobile manipulators additionally require the operator to coordinate locomotion and manipulation simultaneously, and the resulting multimodal sensor streams must still be synchronized and structured into a learning-ready format. We discuss prior work addressing each of these challenges in turn.

\section{Related Work}
\smallskip\noindent\textbf{Cost and scalability of demonstration collection.} Because collecting demonstrations through direct human control is expensive and time-consuming, several methods aim to reduce the amount of operator effort required per demonstration. PATO \cite{Dass-RSS-23} addresses this through policy-assisted teleoperation, automating repetitive parts of the task and requesting operator input only when assistance is needed. MimicGen \cite{mandlekar2023mimicgen} takes a complementary approach by transforming a relatively small set of human demonstrations into a larger and more diverse collection of trajectories. Such methods reduce the amount of direct human control required, but they still depend on an initial collection pipeline capable of reliably recording and organizing human demonstrations.

\smallskip\noindent\textbf{Quality and operator intuitiveness.} Beyond raw collection cost, the design of the teleoperation interface itself strongly affects operator intuitiveness, precision, and ultimately the quality of the collected data. RoboTurk \cite{mandlekar2018roboturk} uses consumer mobile devices to support remote and crowdsourced collection of robot demonstrations. GELLO \cite{wu2024gello} reduces the embodiment mismatch between operator and robot through a low-cost, kinematically matched leader device, while ALOHA \cite{zhao2023learning} employs low-cost leader--follower hardware for precise bimanual teleoperation and imitation learning. The Universal Manipulation Interface instead uses portable hand-held grippers to collect demonstrations outside the target robot workspace \cite{chi2024universal}. Vision and virtual-reality-based systems provide further alternatives: AnyTeleop \cite{Qin-RSS-23} supports different robot arms, hands, camera configurations, and simulated or physical environments within a shared teleoperation framework, while Open-TeleVision  \cite{cheng2025open} provides stereoscopic visual feedback and maps operator motion to the robot for immersive data collection.

\smallskip\noindent\textbf{Coordinating locomotion and manipulation.} Data collection becomes particularly challenging for mobile manipulators because the operator must coordinate navigation and arm control over an extended workspace, rather than manipulation alone. Early work on error-aware imitation learning introduced the MoMaRT interface for collecting demonstrations that combine navigation and manipulation \cite{wong2022error}. TeleMoMa \cite{dass2024telemomamodularversatileteleoperation} provides a modular whole-body teleoperation framework that supports multiple human-input modalities and several mobile-manipulator platforms. Mobile ALOHA \cite{fu2024mobile} uses a dedicated whole-body interface to collect demonstrations for bimanual mobile-manipulation tasks. MoMa-Teleop \cite{honerkamp2025whole} instead delegates mobile-base motion to an assistive policy, allowing the operator to focus on task-relevant end-effector commands. These systems demonstrate different trade-offs between explicit control, simultaneous whole-body operation, specialized hardware, and policy assistance.

\smallskip\noindent\textbf{Structuring multimodal data into learning-ready datasets.} Even once demonstrations are collected, producing a structured dataset from raw sensor recordings remains a substantial engineering effort. Large-scale dataset efforts illustrate the infrastructure this requires: DROID \cite{Khazatsky-RSS-24} uses a standardized data-collection setup to gather manipulation trajectories across many environments, tasks, and operators, while RH20T \cite{fang2024rh20t} records visual, force, audio, action, and robot-state information across a broad collection of manipulation skills and hardware configurations. These datasets demonstrate the value of structured, multimodal robot data, but they also highlight the need for platform-level tools that can acquire, synchronize, inspect, and organize heterogeneous sensor streams before the resulting demonstrations can be used for learning. In practice, sensor streams may run at different frequencies, use different timestamps, and contain corrupted visual frames, while raw recordings often require manual inspection, synchronization, and conversion before use. Operators may also need to switch between separate control interfaces, terminal commands, recording tools, and visualization software, increasing both workload and the risk of inconsistent data.

This paper addresses these four challenges jointly through an integrated teleoperation and multimodal data-collection framework for a Clearpath Dingo-O mobile base equipped with a Kinova Gen3 Lite manipulator and an end-effector-mounted Intel RealSense RGB-D camera. A single DualShock~4 controller is used to operate the base, command the arm in joint-space and Cartesian modes, and start or stop demonstration recording. The controller is coupled to a ROS-based pipeline that records multimodal data, filters selected low-quality image frames, temporally aligns heterogeneous streams, generates replayable rosbag files, and exports the processed demonstrations to the Zarr format \cite{zarr_format}. The focus is not to introduce a new teleoperation paradigm or learning algorithm, but to provide a lightweight end-to-end workflow for collecting structured demonstrations on an existing mobile-manipulation platform.

The framework is validated on a mobile pick-and-place task in which an operator navigates the robot, grasps a bottle, transports it between two surfaces, and releases it at the target location. The experiments demonstrate single-controller operation, automated dataset generation, and synchronized outputs for short-duration demonstrations, while also identifying practical limitations related to network bandwidth, onboard storage, and processing time.

In summary, the main contributions of this work are:
\begin{itemize}
    \item A single-controller interface for base motion, manipulator motion, and demonstration recording;
    \item A ROS-based pipeline for multimodal recording, visual-quality filtering, temporal alignment, replay, and Zarr export; and
    \item A practical validation on a mobile pick-and-place task, including an analysis of deployment limitations on a resource-constrained platform.
\end{itemize}
\section{Approach}
\label{sec:approach}

\subsection{Hardware Architecture and Setup}

The robot used for this project consists of two separate parts. First is the base of the robot, which is a Clearpath Dingo-O \cite{clearpath_dingo}. This is a robotic base that has omnidirectional movement capabilities, which is ideal for moving in dynamic or restrictive environments. The second part of the robot is a Kinova Gen3 Lite \cite{kinova_gen3lite}. This robotic arm has six degrees of freedom and allows objects to be moved accurately in three dimensions.
A RealSense Depth Camera D455 \cite{intel_realsense_d455f} was mounted on top of the end effector to record egocentric image and depth information. This position provides a clearer, more detailed view of the object and allows the camera to be dynamic through space, offering multiple angles to collect data from. A specific mount was designed in a way that it can be attached and detached using a single bolt, allowing quick assembly.

\subsection{Unified Teleoperation}

\begin{figure}[!b]
    \centering
    \includegraphics[width=0.90\linewidth]{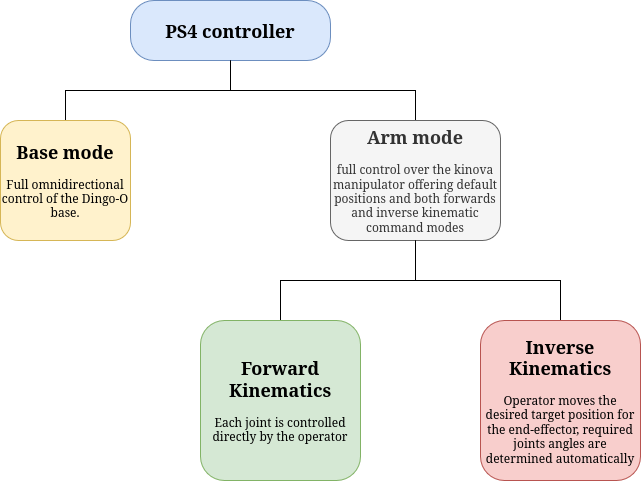}
    \caption{A schematic overview of the different modes implemented in the controller.}
    \label{fig:modeschematic}
\end{figure}

The robot’s teleoperation interface is a crucial factor in ensuring efficient data collection \cite{wan2023teleoperation}. The Dingo-O base and Kinova arm are normally operated using separate DualShock~4 and Xbox One controllers supplied by the manufacturer and vendors. We integrated both into a single DualShock~4 controller connected via Bluetooth. Python-based mode switching lets the same two joysticks control either the base or the arm, as shown in Fig.~\ref{fig:modeschematic}. Base mode retains the manufacturer's controls for forward and sideways motion and rotation about the base's axis.

\begin{figure}[!b]
    \centering
    \begin{tabular}{@{}ccc@{}}
        \includegraphics[width=0.30\columnwidth]{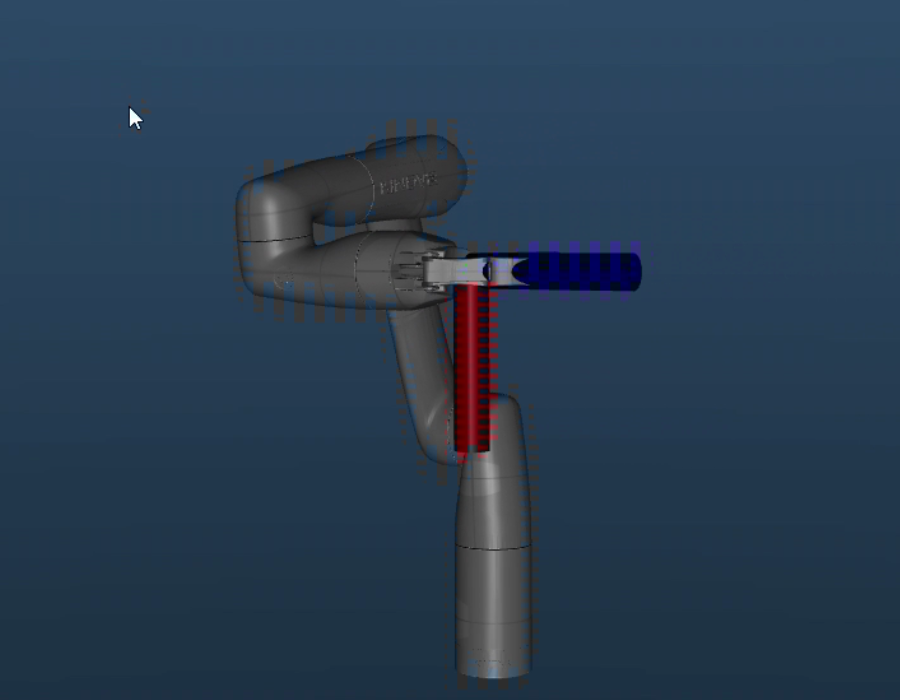} &
        \includegraphics[width=0.30\columnwidth]{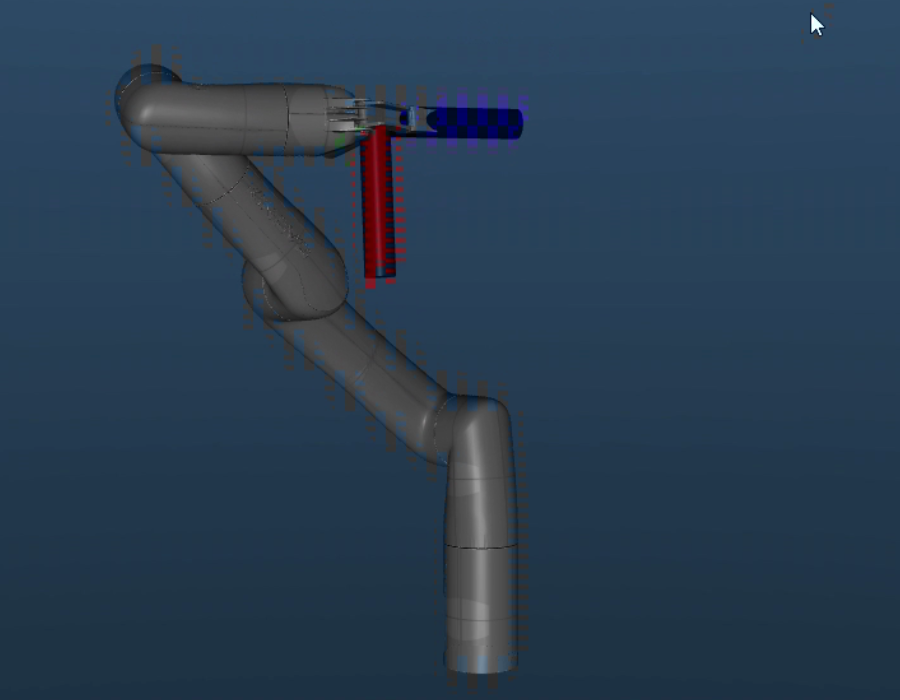} &
        \includegraphics[width=0.30\columnwidth]{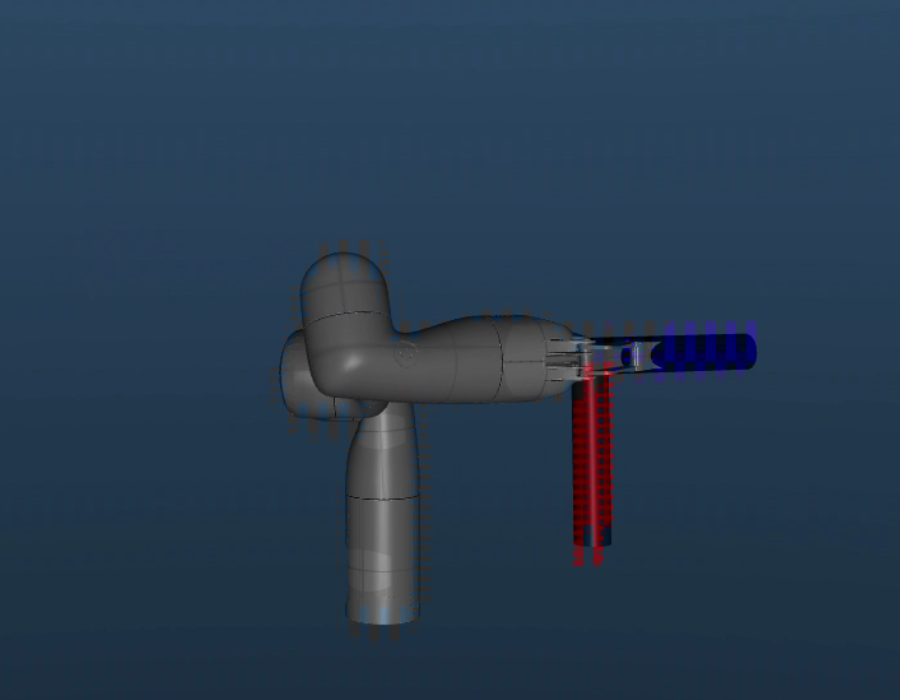}
    \end{tabular}
    \caption{A series of poses of the Kinova arm are shown in the MuJoCo viewer with x, y, and z axes on the end effector.}
    \label{fig:MUJOCO_Sim}
\end{figure}

In arm mode, the same joysticks are used to control the arm’s position, offering both forward and inverse kinematic movement. In forward kinematic mode, each of the six joints of the Kinova arm can be rotated individually, allowing the user to manipulate highly specific joint angles when required. Since the inverse kinematic function requires solving non-linear equations that often have multiple valid solutions, we first built a simulation using MuJoCo (Multi-Joint dynamics with Contact), a widely used physics engine known for its realism \cite{mujoco_simulator}. This was done to adjust gains more effectively if certain movements in the simulation showed any implications. After ensuring smooth and steady joint movements, we transferred the code from the simulation environment to the physical robot to validate real-world functionality, as shown in Figure~\ref{fig:MUJOCO_Sim}.

\begin{figure*}[t]
    \centering
    \begin{tabular}{@{}cc@{}}
        \includegraphics[width=0.45\textwidth]{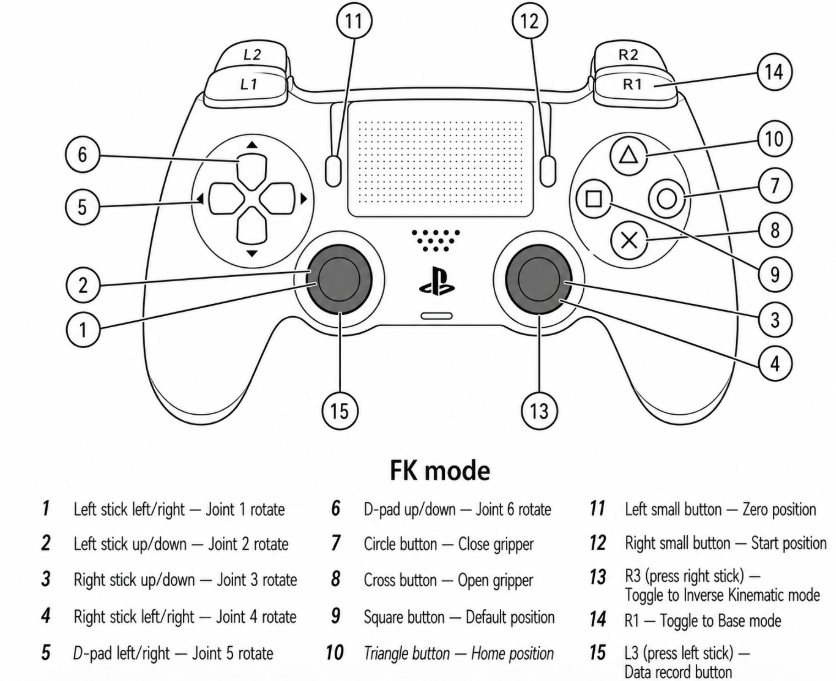} &
        \includegraphics[width=0.45\textwidth]{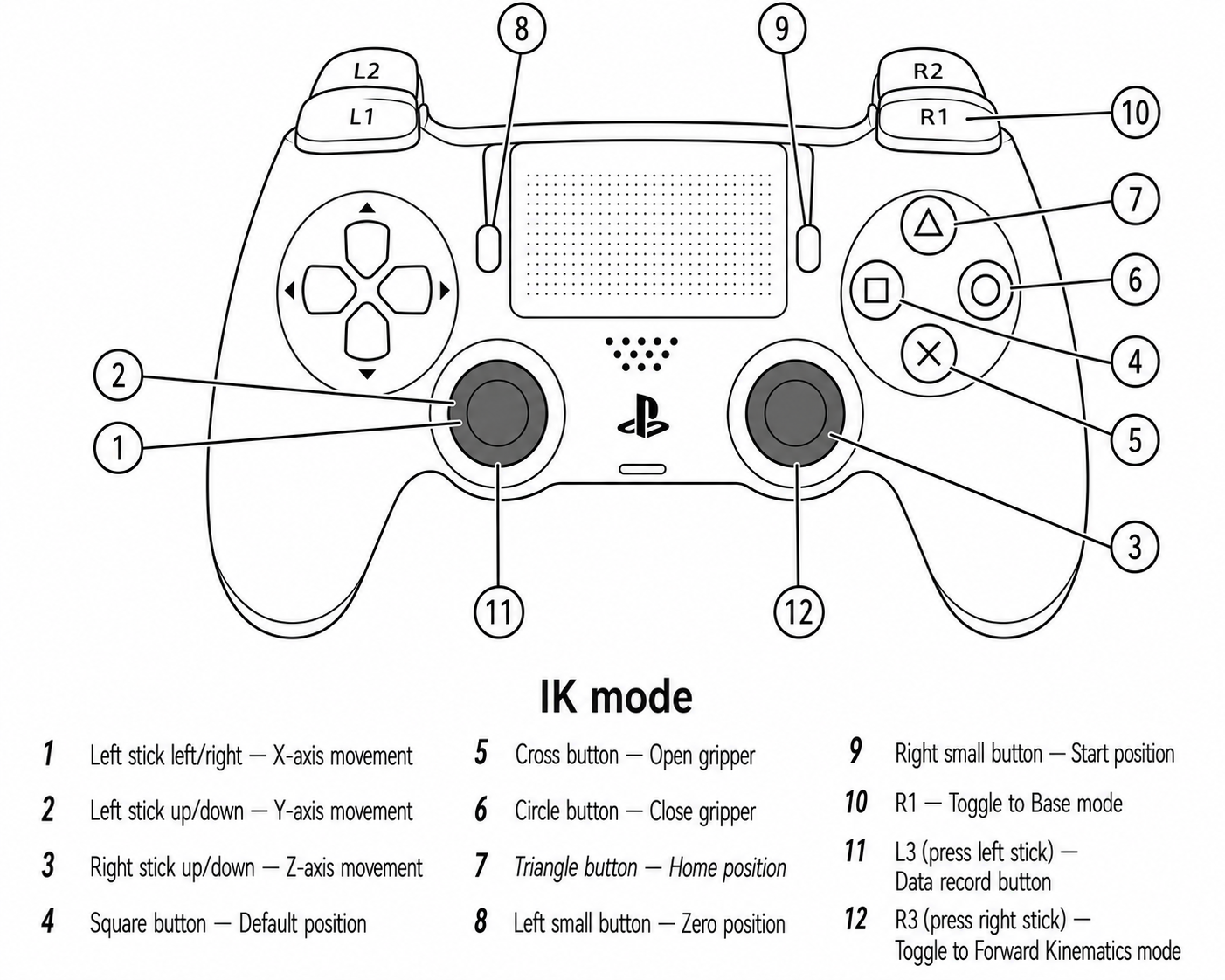} \\
        \scriptsize (a) Forward kinematics &
        \scriptsize (b) Inverse kinematics
    \end{tabular}
    \caption{Controller mappings for the forward kinematics and inverse kinematics modes. The visualization layout was generated using ChatGPT (OpenAI).}
    \label{fig:fk_ik_mapping}
\end{figure*}

Simulation-to-real transfer can be affected by discrepancies in dynamics \cite{peng2018sim}. During initial testing, the physical arm showed jittery movement and inconsistent translation to the desired position. Configuring the Kinova driver to use the manufacturer's high-level controller (HLC) velocity mode during arm teleoperation resolved this issue: the driver correctly interpreted the velocity-based outputs as joint velocity commands, producing smooth and consistent motion.

A controller button starts recording before a task and stops it after completion, triggering an automated Bash script to store and process the demonstration through the pipeline in Section~\ref{sec:data_ingestion}. No external commands are required. Figure~\ref{fig:fk_ik_mapping} shows the full control mappings for the forward and inverse kinematics modes.

\subsection{Data Ingestion, Filtering and Synchronization}
\label{sec:data_ingestion}
The framework was built on Ubuntu 20.04 using ROS1 Noetic for compatibility with other nodes and frameworks developed by the Cognitive Robotics department at TU Delft. Migration to ROS2 is a natural direction for future work. The framework also supports Windows through WSL Ubuntu 20.04 and macOS. An automated Bash script subscribes to active ROS topics, recording wheel encoder data, joint states, end-effector position and raw RGB-D imagery in parallel into a unified rosbag timestamped using the ROS system clock.

The rosbag is saved directly on the host laptop via ROS Network and SSH, reducing onboard storage requirements during small experiments. An automated Python script filters extremely blurry or dark images. Images are compressed and converted to grayscale before the quality check. Using the OpenCV library \cite{opencv_library}, the script compares Laplacian variance and mean grayscale brightness against empirically set thresholds to detect blur and unsuitable brightness. Linear interpolation aligns streams captured at different frequencies, exploiting the robot's smooth motion. The camera serves as the temporal reference because it is central to the learning pipeline and usually has the lowest capture frequency.

\begin{figure*}[t]
    \centering
    \includegraphics[width=0.90\textwidth,trim=0 70 0 60,clip]{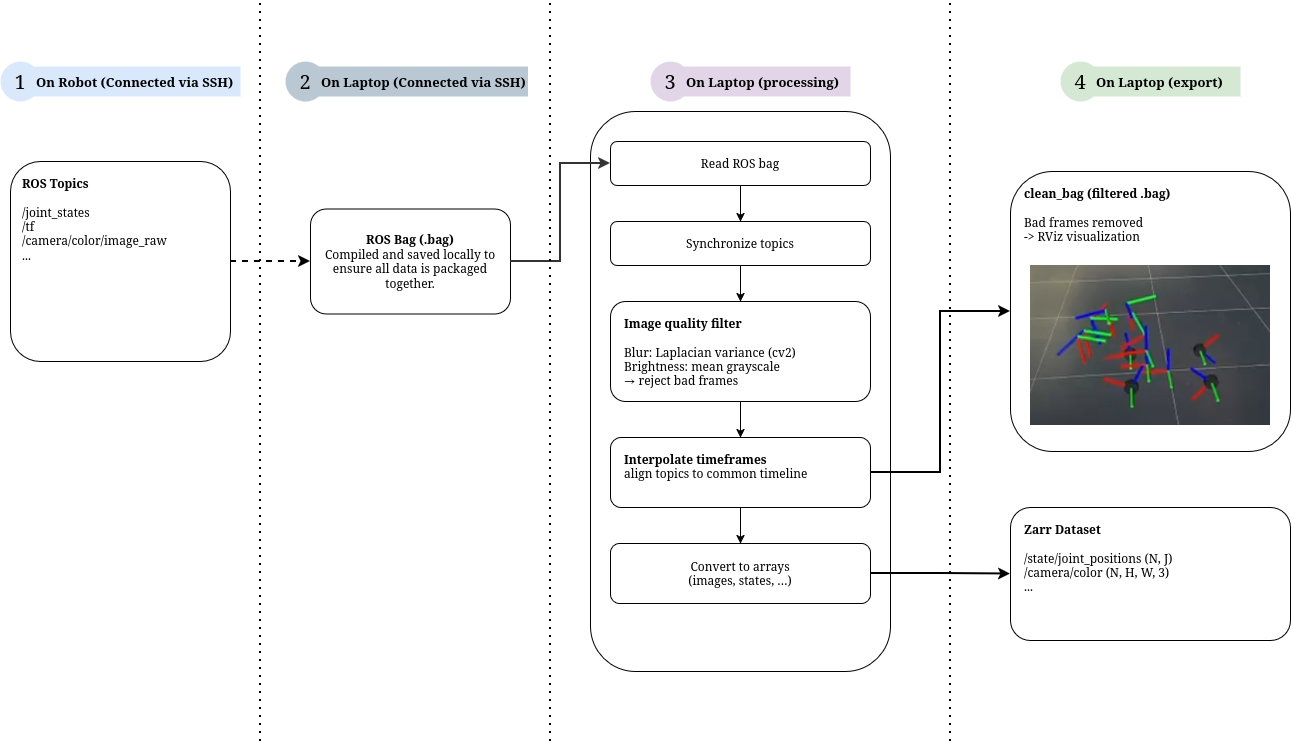}
    \caption{General schematic of the data framework, consisting of data collection, processing and export.}
    \label{fig:dps}
\end{figure*}

Following the quality check and synchronization, the dataset is converted to different formats.
To visualize the synchronized and quality-checked data in RViz, a new .bag file is created.
To efficiently store the data for downstream machine learning pipelines, we utilize the Zarr format \cite{zarr_format}, a flexible and industry-standard architecture for multimodal data storage. This format is suited for storing the synchronized and preprocessed result in a chunked array structure allowing images, joint states, actions, timestamps, and metadata to be stored together in a format that is easy to use in Python-based training pipelines. A general schematic of the data framework is shown in Fig. \ref{fig:dps}.

\section{Experiments and Results}
\label{sec:experiments}
The framework was validated in a mobile pick-and-place experiment that exercised teleoperation, recording, filtering, synchronization, and visualization. The goal was to validate whether a single operator could execute and record a complete demonstration and whether the resulting data could be converted into synchronized outputs for inspection and learning. Code and videos of the experiments can be found at this repo \footnote{Code and videos at \url{https://github.com/sjnsla/SCT-MMDC}}.

\subsection{Experimental setup}
\begin{figure}[htbp]
    \centering
    \begin{tabular}{@{}ccc@{}}
        \includegraphics[width=0.30\columnwidth]{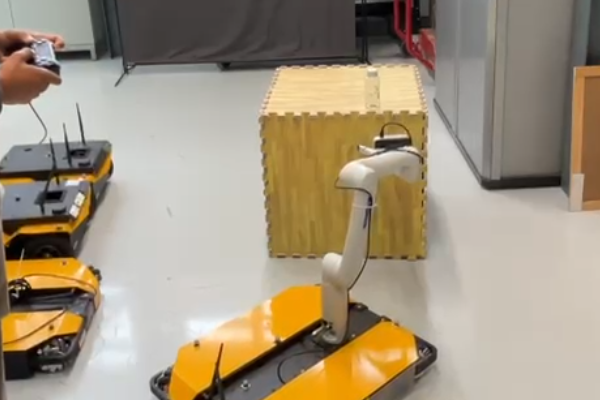} &
        \includegraphics[width=0.30\columnwidth]{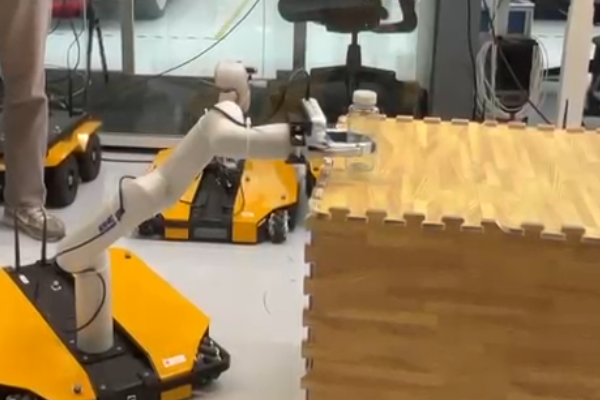} &
        \includegraphics[width=0.30\columnwidth]{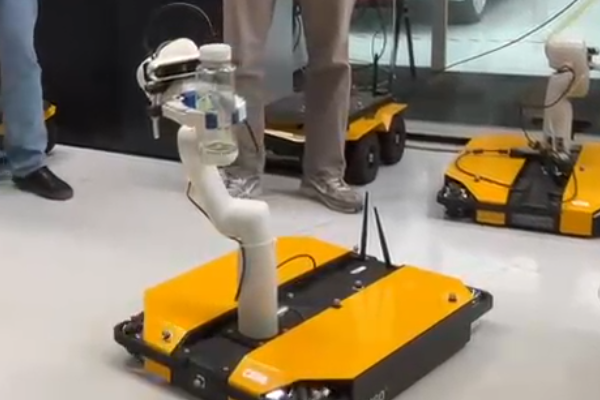} \\
        \scriptsize (a) Navigate &
        \scriptsize (b) Grasp &
        \scriptsize (c) Transport \\
        \includegraphics[width=0.30\columnwidth]{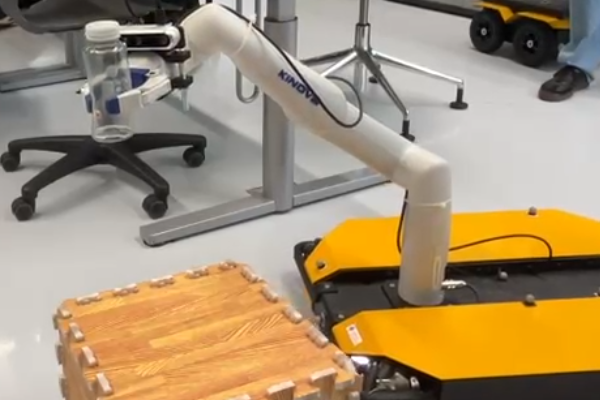} &
        \includegraphics[width=0.30\columnwidth]{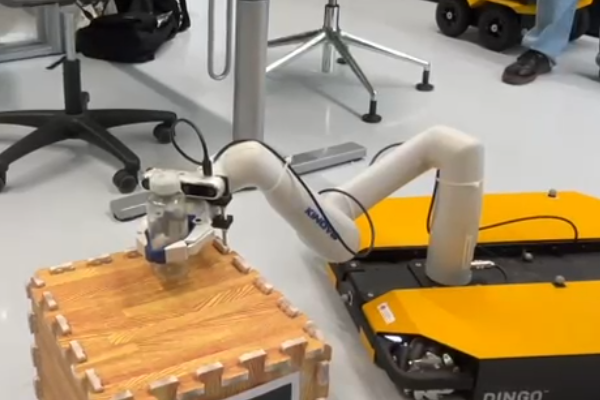} &
        \includegraphics[width=0.30\columnwidth]{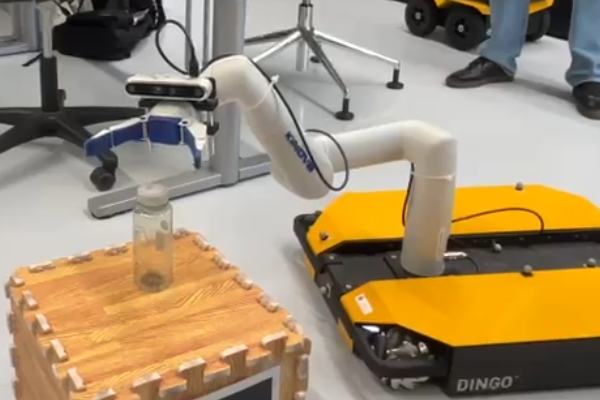} \\
        \scriptsize (d) Stop &
        \scriptsize (e) Release &
        \scriptsize (f) Retreat
    \end{tabular}
    \caption{Experimental sequence of the Dingo-O robot executing the pick-and-place manipulation task.}
    \label{fig:experiment}
\end{figure}

The task required the Dingo-O robot to grasp a bottle, rotate, drive to a lower target surface, and place the bottle upright, as shown in Fig. \ref{fig:experiment}. For this experiment the only required additional objects were an empty bottle and two elevated surfaces, one significantly higher than the other. During each trial, the operator had to press the record button to start the recording of the relevant data. After task completion, the same button stopped the recording and triggered the autonomous processing pipeline, producing both a Zarr dataset and a cleaned rosbag for visualization.  

\subsection{Teleoperation efficiency}
The integration of all controls into the DualShock~4 controller significantly reduced the complexity of operating the robot. The baseline setup required the operator to constantly switch between two separate controllers, one of which required a wired connection, which decreases practicality. During the experiment, the operator successfully managed to repeatedly switch between base mode and arm mode using the assigned button on the controller. This added functionality introduced no significant delay in input response.

Furthermore, the data recording button was also added and validated successfully. When pressing the designated recording button, an SSH command was successfully transmitted via the wireless network to the robot's onboard computer, automatically executing the Bash scripts to initialize ROS bag recording. We observed no delays during the teleoperation and data streaming phase of the experiment.

\subsection{Data Quality filtering \& Synchronization}
\begin{figure}[htbp]
    \centering
    \begin{tabular}{@{}ccccc@{}}
        \includegraphics[width=0.16\columnwidth]{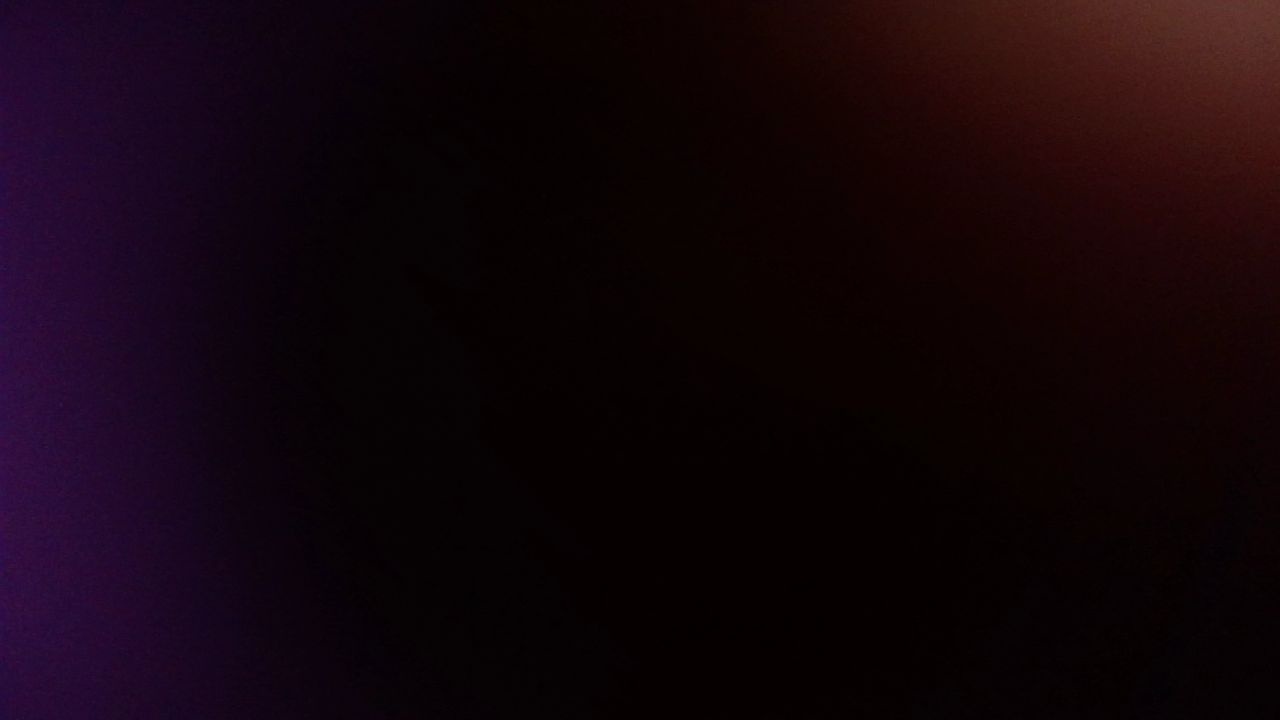} &
        \includegraphics[width=0.16\columnwidth]{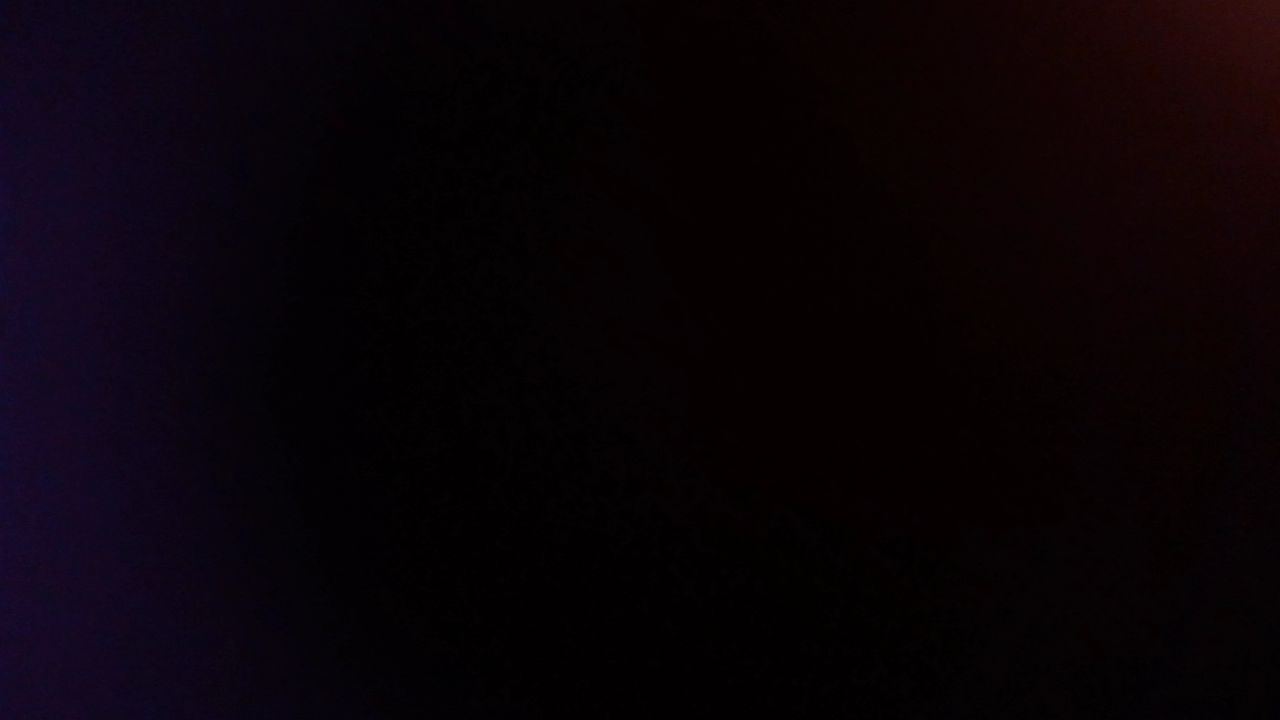} &
        \includegraphics[width=0.16\columnwidth]{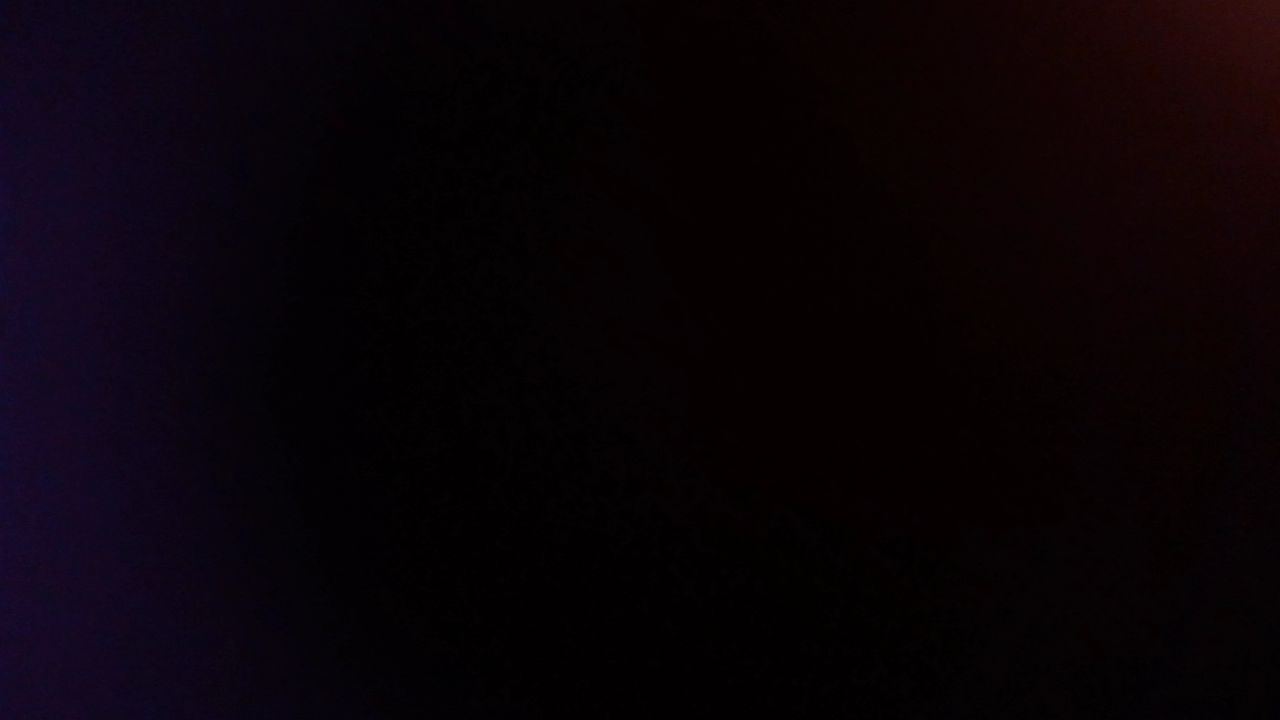} &
        \includegraphics[width=0.16\columnwidth]{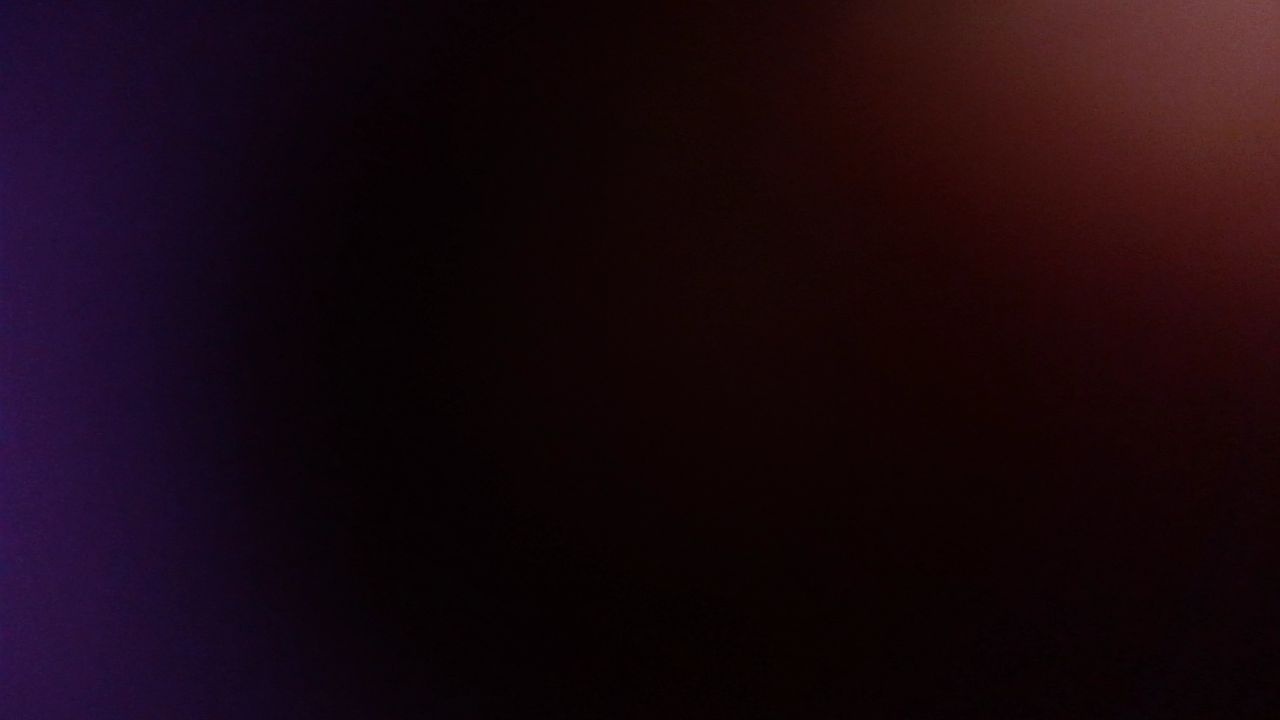}
    \end{tabular}
    \caption{Examples of overly dark visual frames rejected by the quality filter.}
    \label{fig:blur}
\end{figure}

The recorded data were processed using the filtering and synchronization pipeline described in Section~\ref{sec:data_ingestion}. This was done successfully and resulted in several images of unusable quality being filtered out, some of which are shown in Fig. \ref{fig:blur}. Across five trials, the filter rejected on average 1 out of every 300 frames at 30 frames per second. Manual verification of both the approved and rejected image frames confirmed the precision of the filter. The thresholds for the Laplacian variance and grayscale mean brightness were systematically calibrated to precisely identify and discard corrupted visual frames while minimizing false positives.

Regarding the temporal synchronization, the linear interpolation method proved to be computationally efficient and accurate in synchronizing the high-frequency data from the encoders with the lower-frequency data from the camera. The camera feed successfully served as the reference for the temporal alignment.

\subsection{Framework validation \& visualization}
\begin{figure}[htbp]
    \centering
    \begin{tabular}{@{}cc@{}}
        \includegraphics[width=0.45\columnwidth]{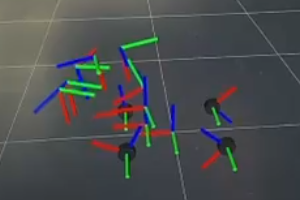} &
        \includegraphics[width=0.45\columnwidth]{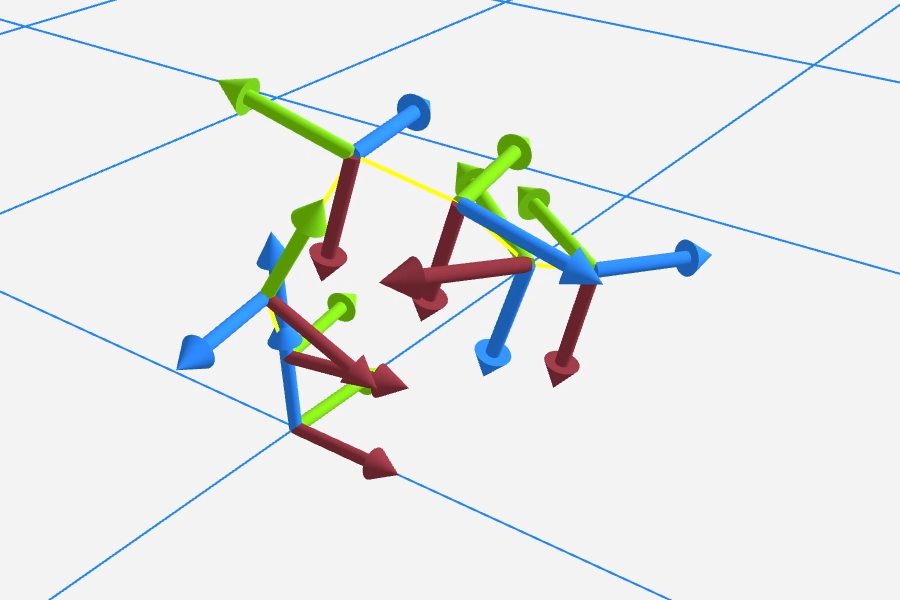} \\
        \scriptsize (a) RViz &
        \scriptsize (b) Foxglove Studio
    \end{tabular}
    \caption{Comparison of RViz and Foxglove Studio for visual inspection of the synchronized demonstration.}
    \label{fig:rvfox}
\end{figure}

The cleaned rosbag output was evaluated using RViz and Foxglove Studio \cite{foxglove_platform}. As described in Section~\ref{sec:data_ingestion}, the pipeline stores a copy of the cleaned data before conversion to Zarr and reformats this copy as a rosbag for visualization. Although Zarr is the primary output format for learning, the cleaned rosbag allows the robot state to be replayed together with the synchronized camera frames. This provides a practical inspection step for checking whether the reconstructed manipulator motion is consistent with the recorded images. In our setup, Foxglove Studio provided clearer visual feedback than RViz for the manipulator model because mesh-rendering issues in RViz left mainly the joint axes visible, as shown in Fig.~\ref{fig:rvfox}. This limitation means that synchronization quality is still assessed visually; future versions should include quantitative timing checks against external measurements or known event markers.

The experiments successfully validated the main functionality of the proposed teleoperation and data-collection framework. Four users without previous training were able to complete the pick-and-place task and record demonstrations without encountering major difficulties. This indicates that the controller mapping and recording procedure can be used by new operators with limited preparation. Nevertheless, we acknowledge that the experimental evaluation was limited and should not be interpreted as a formal usability assessment. Its purpose was to verify that the integrated system operated correctly and that multiple users could collect demonstrations. A larger user study measuring task-completion time, error rates, learning effort, and operator workload is therefore left for future work.

The selected experiment imposed relatively modest requirements on the platform. It did not involve high-speed motion, precise object placement, or rapid coordination between the mobile base and manipulator. Consequently, the reliability of the framework under high-speed or high-precision operating conditions has not been established. Additional cameras and sensing modalities can also be incorporated in future versions of the framework.

Alternative input devices, such as a SpaceMouse and virtual-reality motion controllers, were also considered because they can provide more direct control over higher-dimensional motion. The DualShock~4 controller was selected because of accessibility, cost and familiarity. Previous work has also shown that gamepad teleoperation can offer competitive performance compared with more specialized teaching interfaces in industrial settings \cite{Massa2025GamepadTeleoperation}. The teleoperation software is not restricted to the DualShock~4, as its Python-based control mappings can be reconfigured to support other input devices.

The processing pipeline operated reliably during short-duration experiments, consistently producing synchronized Zarr datasets and cleaned rosbag files. Longer recordings, however, revealed limitations related to processing time, storage, and network communication. These limitations could be reduced by executing the complete recording and processing pipeline onboard the robot. Adding dedicated solid-state storage would support longer recordings and remove the need to continuously stream raw data over the network. Only the final Zarr dataset and cleaned rosbag would then need to be transferred to the host laptop. Overall, the current experiments validate the feasibility of the integrated workflow, while broader usability studies and more demanding manipulation tasks remain necessary to evaluate its performance more comprehensively.
\section{Conclusion}
\label{sec:conclusion}

This paper presented an integrated framework for teleoperated multimodal data collection with a Dingo-O mobile base, Kinova Gen3 Lite manipulator, and end-effector-mounted RealSense RGB-D camera. The system combines single-controller operation of the base, arm, and recording process with an automated ROS-based pipeline for recording, image-quality filtering, temporal synchronization, visualization, and export to Zarr. In a mobile pick-and-place experiment, the unified controller enabled a single operator to execute and record complete demonstrations without external commands, while the data pipeline produced synchronized outputs for short-duration trials and rejected visually corrupted frames.

The experiments also revealed practical limitations that must be addressed before scaling to longer recordings or larger datasets. A natural next step is to run the full pipeline onboard the robot after adding dedicated solid-state storage, transferring only the final Zarr and cleaned rosbag files after processing. Future work should also include large-scale user studies that assess the framework and incorporate user feedback in the recording and operation process.

\bibliographystyle{IEEEtran}
\bibliography{references}

\end{document}